# VIDEO SUMMARIZATION: STUDY OF VARIOUS TECHNIQUES

**[1]RAVI RAJ, [2]VARAD BHATNAGAR, [3]AMAN KUMAR SINGH, [4]SNEHA MANE, [5]NILIMA WALDE**

[1,2,3,4,5] Department of Information Technology, Army Institute Of Technology, Pune, India
E-mail: varadhbhatnagar@gmail.com

**Abstract -** A comparative study of various techniques which can be used for summarization of Videos i.e. Video to Video conversion is presented along with respective architec- ture, results, strengths and shortcomings. In all approaches, a lengthy video is converted into a shorter video which aims to capture all important events that are present in the original video. The definition of 'im- portant event' may vary according to the context, such as a sports video and a documentary may have different events which are classified as important.

**Keywords -** Video Summarization, DNN, RNN, Deep Semantic Features, Confusion Matrix

## I. INTRODUCTION

Due to the rapid growth of camera technology and internet affordability, web applications offering video host- ing services have increased rapidly. Youtube, Vimeo, Hotstar, Dailymotion to name a few. A large number of videos spanning varied genres such as sports, enter- tainment, movie, news coverage, etc are uploaded on the internet daily, so it becomes a difficult task for a user to choose a relevant video which he/she may want. There are many video retrieval techniques available most of which depend on user description, tags and thumb- nails which are not very descriptive. It also depends on the content creator who can add click-baits and irrelevant description which are not accurate.

To address these challenges one solution can be automatic video summarization. This will provide a insightful summary of the large videos which will help user effectively find videos on web within short period of time.

The Proposed technique in this paper uses deep se- mantic features to summarize videos. We divide original video into smaller segments and for each segment we calculate deep features in high dimensional, continuous space using DNN.Then semantic features which are not redundant are selected such that combination of which will represent the entire video. Then all the sampled seg- ments are concatenated to make video summary.

## II. LITERATURE SURVEY

### A. Video Summarization with Long Short-Term Memory [1]

Video Summarization with Long Short-Term Memory is a technique which has been used for efficient summa- rization of videos. It is a Supervised Learning Technique which uses user annotated videos as its input. These videos have been seen by users, frame by frame and an- notated according to their understanding of the events unfolding and the storyline. The user summary may vary because it is heavily subjective and can be biased.

For accurate summary generation of a video, we need to understand that there exists a Temporal Dependency between the frames of a video. Temporal Dependency means that the frames are dependent on one another in a variety of ways such as time of the day, progression of storyline, etc.

The concept of Structured Predictions is crucial to understand this approach. The problem of Video Summarization can be solved by structured prediction which is very different from binary or real valued prediction. Ac- cording to this concept, the frames of the video cannot be treated in an isolated manner. It is not a set but a sequence which needs to be analysed to get a feasible summary covering all highlights.

There are two ways to summarise a video. The first can be Keyframe Selection. In this, only the frame num- bers of important highlights are output by the model. The other method can be Key Subset Selection, in which subsequent ranges are present in the output in which the highlights reside.

LSTM, a type of Recursive Neural Network is used in this approach because memory is required to remember long range and short range interdependencies between the frames which cannot be ignored. The memory ele- ment and the control gates model this behaviour. A basic illustration of LSTM is given in the below diagram which shows the layout of the memory element and the input, output and the forget gates.

The authors of this paper have designed a specific model for this purpose which they have called vsLSTM (Video Summarization Long Short Term Memory). They have added a few features to the basic model such as bidi- rectional LSTM Layers to model long range dependency in both past and future directions. It outputs a score associated with each frame known as Frame Level Importance Score.

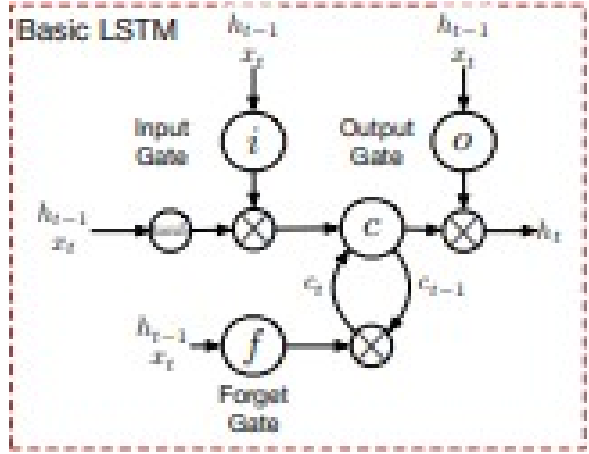

$$i_t = \text{sigmoid}(W_i[x_t^T, h_{t-1}^T]^T)$$
$$f_t = \text{sigmoid}(W_f[x_t^T, h_{t-1}^T]^T)$$
$$o_t = \text{sigmoid}(W_o[x_t^T, h_{t-1}^T]^T) \quad (1)$$
$$c_t = i_t \odot \tanh(W_c[x_t^T, h_{t-1}^T]^T) + f_t \odot c_{t-1}$$
$$h_t = o_t \odot \tanh(c_t),$$

A mathematical approach called Determinantal Point Process (DPP) is used in combination with the LSTM. It selects a subset of frames keeping the diversity of frames (summarization structure) into consideration. The re- sults of this approach can be evaluated by the following two measures:

**Precision**: Overlapped Duration of A and B / Duration of A

**Recall**: Overlapped Duration of A and B / Duration of B

| Dataset | Method | Canonical | Augmented | Transfer |
|---|---|---|---|---|
| SumMe | MLP-Shot | **39.8**±0.7 | 40.7±0.7 | 39.8±0.6 |
| | MLP-Frame | 38.2±0.8 | 41.2±0.8 | 40.2±0.9 |
| | vsLSTM | 37.6±0.8 | 41.6±0.5 | 40.7±0.6 |
| TVSum | MLP-Shot | **55.2**±0.5 | 56.7±0.5 | 55.5±0.5 |
| | MLP-Frame | 53.7±0.7 | 56.1±0.7 | 55.3±0.6 |
| | vsLSTM | 54.2±0.7 | 57.9±0.5 | 56.9±0.5 |

The limitations of this technique are that the training data set needs to be large and diverse. The frame by frame Annotation of videos by a human is a very slow, repetitive and mundane task. Without a strong dataset in which videos are annotated in detail, this technique will fail to meet accuracy standards. Another limita- tion to this technique is that it is giving good results for videos in which the content is changing in a smooth manner as opposed to Videos with rapidly changing con- tent and High Level Semantic Features. For obtaining high accuracy in such videos, the authors believe that another technique such as object detection will need to be incorporated into this.

Where,
A: Keyshot based Summary B: User Summary

**B. Fast-forward Video Based On Semantic Extraction [2]**
Fast-forward Video Based On Semantic Extraction technique extracts semantic information and further it is encoded with score function which has 3 following components : 1. Confidence of extracted information 2. Centrality of the analysed region 3. Size of the region

Score for the frame i and K detected Regions of Interest (ROI) is given by:- Si = C(k) G(k) A(k) Semantic Amount is the semantic information present in the final video.

Jitter amount is measured between consecutive output frames through the mean magnitude of the FOE locations differentiation. The lower is the amount of jitter the smoother is the video.

Speed-up deviation is the distance between desired and and achieved speed-up technique.

Image processing and semantic analysis is used to divide the video into semantic and non- semantic segments and semantic scores are calculated. Gaussian function is used to remove outliers and mean value between maxi- mum and minimum value is used to define threshold and segments above this threshold are classified as semantic and remaining are classified as non-semantic.

Fs values the segments classified as semantic and Fns values are classified into non-semantic. Optimization function is used to select Fs and Fns. Speed-up rates are estimated based on length of each part one for seman- tic and the other for the non-semantic part. A graph is produced from each part and minimization is computed using shortest path algorithm to create final video, se- lected frames of each segment compose the final video

**C. Automatic Summarization of Soccer Highlights Using Audio-visual Descriptors [3]**
Automatic highlights summarization generation of soccer videos using audio-visual descriptors is used to generate automatic summary of soccer game. This technique is based on sport video edition human-expertise used in commercial television. Objective is to divide input soccer video into shots and each shot is assigned with a score and then shot with highest score value is selected to build up the summary. Shot can be defined as the minimum unit for building up the summary which has essential events of interest.
High level audio-visual descriptor is used to annotate these shots. In this approach shot boundary is detected using abrupt shot detector during action of the game and the dissolve detector, along with a whistle detector algorithm. Keyframe extraction within a shot is done by processing all the motion compensation vectors within a shot and performs a full search analysis in order to find first the frame with the maximum motion activity. Depending on the intensity of action it summarizes each shot in keyframe.

Low-level Descriptors, Persons Descriptor, Replay Detector, Zoom Detector, Long Shot Detector, Whistle Detector, Inter and Intra shot-based audio detectors are used to produce to generate the video soccer match summarization when combined with other descriptors which is passed through highlight generator to generate the final summary.
The Limitation of this technique is it produces false detection because of too much cheering noise nearby the whistles frequency band.[7]

**D. Highlight Summarization In Soccer Video Based On Goalmouth Detection [4]**
Highlight Summarization in Soccer Video based on Goalmouth Detection is a technique that has been used for summarization of football matches. It is a Supervised Learning Technique which uses Match Video as input along with the audio. The match videos that have been used to train and test the model are from the Olympic 2004 and Eurocup 2004 events. The basic approach is divided into two parts. The first part is to classify all frames present in the video in to Goalmouth and Non Goalmouth frames[6]. The region surrounding the Goalpost is referred to, as Goalmouth through the paper. The authors have assumed that all exciting events in a match happen near the Goalmouth and hence it should be the targeted area. The frames are classified using a SVM classifier and features such as:

1. Slope of Top Boundary
2. Slope of Bottom Boundary
3. Slope of Left Boundary
4. Slope of Right Boundary
5. Corner Positions

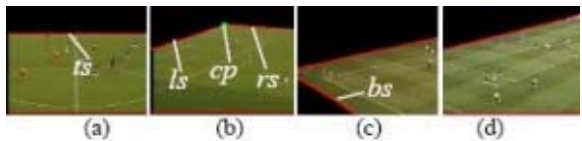

Figure 5: Shape features of a frame

After classification of the frames, the next part is the ranking of frames. This is done keeping the excitement level of the audio in mind. It is a common phenomenon for the crowd and the commentators to get more excited and applaud in the event of a goal. The authors are harnessing this feature in ranking the highlights. Then based on a threshold value, the frames which meet the criteria are concatenated to form the highlight video as shown in the picture.

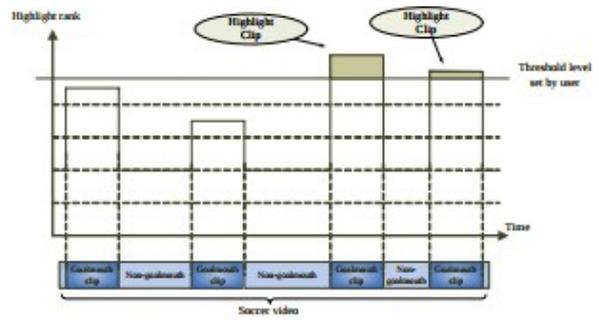

Figure 6: Goalmouth and non-goalmouth clips in a video

The results of this approach are given as per the following matrix:
There are a couple of limitations which have been high-lighted by the authors, the first and foremost being the camera angles. Since, the coverage of a match depends on the broadcaster, the goalmouth area may get captured differently in different type of matches. The other limitation of this approach is that there are a lot of commentators present who officiate matches and their style is very different. So, it is a problem to generalize over all types of commentary styles.

|  | Precision | Recall |
|---|---|---|
| EuroCup 2004 | 87.2% | 73.2% |
| Olympic 2004 | 86.7% | 81.3% |

Table 1: Experiment of the goalmouth frames detection

**III. OUR INTENDED APPROACH**

In this approach we extract uniform length video segments from an input video. The segments are then given to CNN which extracts features and it is then mapped to points in a semantic space. Video summary is then generated by sampling video segments that correspond to cluster centers in the semantic space.[8]

We are using the Soccernet Dataset by Silvio Giancola to train our model[5]. This data-set comprises of more than 700 hours of football match footage with labelled frames depicting events.

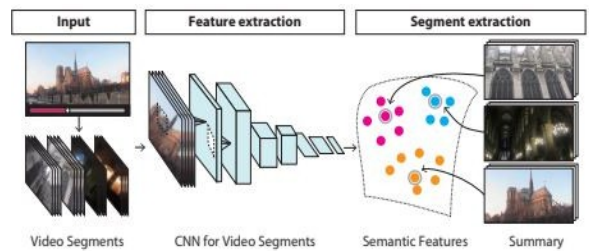

To increase the efficiency we use DNN of two sub net- works video and its description are mapped to a com- mon semantic space. Video sub network uses CNN and description sub network uses recurrent neural network (RNN). Contrastive loss function is used during train- ing to reduce the gap between relevant description and video and irrelevant sentence and video are moved apart. Contrastive loss function is defined as :- loss(Xn, Yn) = tnd(Xn, Yn) + (1 tn) max(0, d(Xn, Yn)) Where Xn,Yn are extracted features for the n-th video and description pair, d(Xn, Yn) is the squared Euclidean distance between Xn and Yn, if (Xn, Yn) is positive then tn = 1, otherwise tn = 0 .

### A. Learning Deep Features

Our video sub-network consists of two fully-connected layers with hyperbolic tangent followed by a mean pool- ing layer which fuses different frames in a video segment. Video segment is divided into frames and it is given frame by frame to the video sub-network to compute video representation. Simultaneously description sub-network use skip-thought vector to encode the sentence into 4800 dimensional vector using RNN. These two sub-networks are trained jointly to compute negative and positive pairs.

### B. Generating Video Summary

k-medoids algorithm is a clustering algorithm related to the k-means algorithm and the medoid shift algorithm which attempts to minimize the distance between points labeled to be in a cluster and center of that cluster this technique is used to generate video summary. k-medoids is used to find subset S X which minimises objective function where S = Sk—k = 1, . . . , K is a subset of cluster centers of video segment, K is a given parameter to determine the length of the video summary and X = Xj —j = 1, . . . , L is the set of deep features extracted from all video segments in the input video, The optimal subset S = argmin S F(S) is calculated which contains the most representative and diverse video segments in clus- ters. The segments in S are concatenated in the temporal order to generate a video summary.

## FUTURE DIRECTION

This idea holds a lot of application in the world of sports where the need for highlights of matches is required. It is still being done manually and such repetitive work could be easily automated by using a fully developed and error resistant model. The number of videos on a specific topic are bound to increase in the coming years. Effective filtering of these videos to find most appropriate can also be done by this.

## CONCLUSION

In this paper we proposed a novel approach to retrieve a large video with the help of deep neural network (DNN). Our method uses RNN and CNN in sentence sub-network and video sub-network respectively and train them jointly using contrastive loss function to map positive segments close to the description in the se- mantic space making a cluster. In our approach, the in- put video is represented by deep features in the semantic space, and segments corresponding to cluster centers are extracted to generate a video summary. By comparing our summaries to manually created summaries, we shown that the advantage of incorporating our deep features in a video summarization technique.

## ACKNOWLEDGMENTS

We would like to thank our Head of Department, Dr Sangeeta Jadhav for giving us the opportunity to work on this project and paper. We would like to thank our project guide Prof. Nilima Walde for providing us key insights and valuable suggestions as we went about com- pleting this paper. Lastly, we would like to thank each other for good teamwork.

★ ★ ★